\documentclass{article}

\usepackage[preprint]{neurips_2024}

\usepackage[utf8]{inputenc} 
\usepackage[T1]{fontenc}    
\usepackage{hyperref}       
\usepackage{url}            
\usepackage{booktabs}       
\usepackage{amsfonts}       
\usepackage{nicefrac}       
\usepackage{microtype}      
\usepackage{xcolor}         

\usepackage{CJKutf8}
\usepackage{tikz}
\usetikzlibrary{shapes.geometric}
\usetikzlibrary{arrows.meta,arrows,calc}
\usepackage{enumitem}
\usepackage{stackengine}
\usepackage{multirow}

\usepackage{hhline}

\title{The Breeze 2 Herd of Models: Traditional Chinese LLMs Based on Llama with Vision-Aware and Function-Calling Capabilities}

\author{%
  Breeze Team @ MediaTek Research\thanks{
  All contributors to this work have made equal contributions. We list all contributors in alphabetical order by first name, as follows: Chan-Jan Hsu, Chia-Sheng Liu, Meng-Hsi Chen, Muxi Chen, Po-Chun Hsu, Yi-Chang Chen, and the supervisor Da-Shan Shiu. The author contributions are detailed in the end of this document.
  }
}

\begin{document}

\maketitle

\begin{abstract}
\textit{Llama-Breeze2} (hereinafter referred to as \textit{Breeze2}) is a suite of advanced multi-modal language models, available in 3B and 8B parameter configurations, specifically designed to enhance Traditional Chinese language representation. Building upon the Llama 3.2 model family, we continue the pre-training of \textit{Breeze2} on an extensive corpus to enhance the linguistic and cultural heritage of Traditional Chinese. In addition to language modeling capabilities, we significantly augment the models with function calling and vision understanding capabilities. At the time of this publication, as far as we are aware, absent reasoning-inducing prompts, \textit{Breeze2 } are the strongest performing models in Traditional Chinese function calling and image understanding in its size class. The effectiveness of \textit{Breeze2} is benchmarked across various tasks, including Taiwan general knowledge, instruction-following, long context, function calling, and vision understanding. We are publicly releasing all \textit{Breeze2} models under the Llama 3.2 Community License\footnote{\url{https://huggingface.co/collections/MediaTek-Research/breeze-2-67863158443a06a72dd29900}}. We also showcase the capabilities of the model running on mobile platform with a mobile application which we also open source\footnote{\url{https://github.com/mtkresearch/MR-Models}}. 
\end{abstract}

\section{Indroduction}

In recent years, the advent of foundation models has revolutionized the field of artificial intelligence, enabling a wide array of tasks across different modalities such as language and vision. These models, which undergo extensive pre-training and post-training stages, have demonstrated remarkable capabilities in understanding and generating natural language. However, a significant gap remains in the representation and performance of these models concerning Traditional Chinese, a language predominantly used in Taiwan, Hong Kong, and Macau.

Traditional Chinese, with its rich linguistic and cultural heritage, has been underrepresented in the development of large language models (LLMs). Prominent models such as GPT-4o\footnote{\url{https://openai.com/index/hello-gpt-4o/}}, Claude 3\footnote{\url{https://www.anthropic.com/news/claude-3-family}}, and Llama 3 \citep{dubey2024llama} have shown exceptional proficiency in English and Simplified Chinese. Nevertheless, they often fall short in accurately capturing the nuances and complexities inherent in Traditional Chinese. This misalignment not only affects the accuracy of language generation but also overlooks the cultural significance embedded within Traditional Chinese texts.

While previous works \citep{hsu2024breeze,lin2023taiwan} have attempted to address this gap, they lack vision-aware and function-calling capabilities, both of which are crucial for commercial applications. Vision-aware capabilities enable models to interpret and generate content that integrates text and images, essential for industries like e-commerce and virtual assistants. Function-calling capabilities allow models to interact with external systems, perform specific tasks, and provide accurate responses, enhancing applications like automated customer support.

This work aims to address the underrepresentation of Traditional Chinese in LLMs by developing a model that integrates vision-aware and function-calling capabilities. The result of our efforts is \textit{Breeze2}: a suite of two multi-modal language models with 3B and 8B parameters (Section \ref{sec:architecture}). \textit{Breeze2} is built upon the impressive foundation of Llama 3 \citep{dubey2024llama} and continued pre-trained on 900 GB of high-quality Traditional Chinese corpus (Section \ref{sec:pretrain-data-loc}) to enhance the linguistic and cultural heritage of Traditional Chinese. To enable vision-aware capabilities, we adapted the design from LLaVA \citep{liu2023llava} by integrating a visual encoder and a bridge module, and employed a two-stage training process: vision alignment and visual instruction tuning (Section \ref{sec:method}). To enable function-calling capabilities, we adapted the design from Breeze-FC \citep{chen2024enhancing} by integrating prompt templates for function calling (Section \ref{sec:prompt-template}) and conducted post-training on function-calling data (Section \ref{sec:post-data-fc}).

To comprehensively evaluate the effectiveness of \textit{Breeze2}, we conducted extensive benchmarking across a wide range of tasks, including Taiwan general knowledge (Section \ref{sec:benchmark-tw}), instruction-following (Section \ref{sec:benchmark-tw}), long context handling (Section \ref{sec:benchmark-long}), function calling (Section \ref{sec:benchmark-fc}), and vision understanding (Section \ref{sec:benchmark-vis}). Additionally, the 3B parameter model is designed to be more adaptable for mobile use, and we demonstrate \textit{Breeze2}'s capabilities in a mobile application (Section \ref{sec:app}).

\section{Model Architecture}
\label{sec:architecture}

Building upon our goals to address the underrepresentation of Traditional Chinese in LLMs, \textit{Breeze2} adopts the widely-used ``ViT-MLP-LLM'' architecture, originally introduced by LLaVA \citep{liu2023llava}, which has become a common configuration in open-source multimodal large language models. This architectural choice allows \textit{Breeze2} to effectively leverage the strengths of both vision and language models, while also incorporating improvements inspired by subsequent works such as InternVL 1.5 \citep{chen2024fargpt4vclosinggap}.

\begin{figure}[htbp]
\centering
    \includegraphics[width=0.7\textwidth]{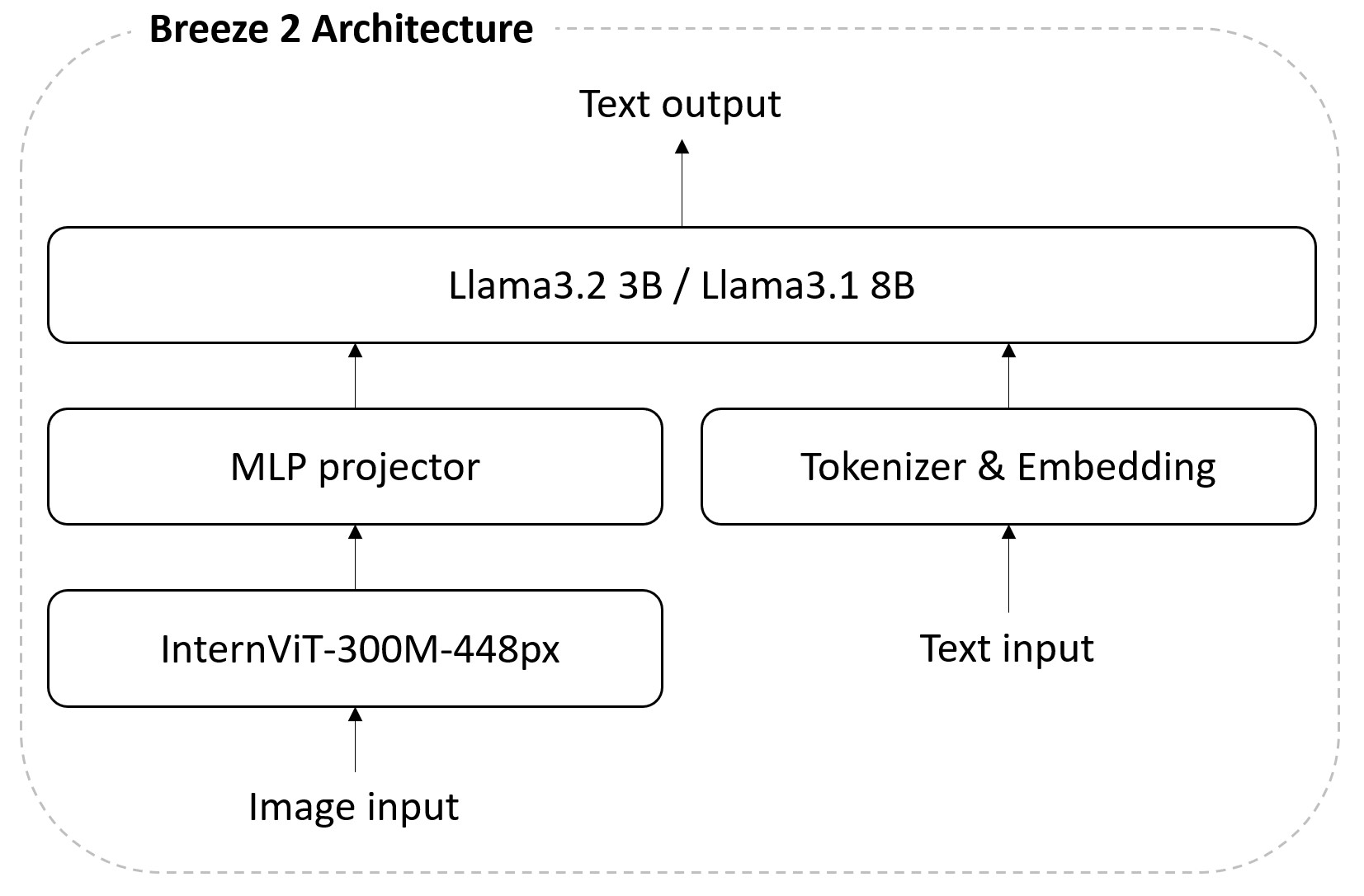}
    \caption{\textit{Breeze2} Architecture. \textit{Breeze2} utilizes the widely-adopted “ViT-MLP-LLM” paradigm, which integrates a pre-trained InternViT-300M-448px with LLMs of various sizes through a vision-to-language MLP projector. Specifically, it combines the InternViT-300M-448px with Llama 3.2 3B and Llama 3.1 8B models. This architecture leverages the strengths of ViT, MLP, and LLM to process both visual and textual inputs effectively.}
    \label{fig:breeze2_architecture}
\end{figure}

As illustrated in \autoref{fig:breeze2_architecture}, the implementation of this architecture in \textit{Breeze2} involves the integration of several key components:

\begin{itemize}
\item A pre-trained Llama 3 model, serving as the foundation for language understanding and generation. This includes its pre-trained tokenizer and embedding layer for processing text inputs.
\item A pre-trained InternViT-300M-448px model, which acts as the vision encoder.
\item A vision-to-language projector, implemented as a multi-layer perceptron (MLP), which serves as a bridge between the visual and linguistic features.
\end{itemize}

These components work together to process both visual and textual inputs effectively.

The choice of the InternViT-300M-448px model as the vision encoder was made for two significant reasons:

\begin{enumerate}
\item \textbf{Visual Processing Capabilities:} The 448px input resolution allows for detailed image analysis, which is crucial for tasks requiring fine-grained visual understanding.

\item \textbf{Enhanced Chinese Language Understanding:} Notably, the InternVL pretrained model demonstrates a higher level of comprehension for Chinese characters compared to other vision models. This enhanced capability in processing Chinese text within images gives \textit{Breeze2} a distinct advantage in Traditional-Chinese-specific visual-language tasks.
\end{enumerate}

\textit{Breeze2} is available in two model sizes, each tailored for different use cases:

\begin{itemize}
\item \textbf{Breeze2 3B:} This variant is based on the Llama 3.2 3B model, offering a balance between performance and computational efficiency, making it suitable for resource-constrained environments.
\item \textbf{Breeze2 8B:} This larger variant is based on the Llama 3.1 8B model, providing enhanced capabilities for more complex tasks that require higher capacity and performance.
\end{itemize}

The choice of these specific Llama 3 variants as the base models for \textit{Breeze2} allows for a range of applications, from more resource-constrained environments to scenarios requiring higher capacity and performance.

The vision-to-language projector plays a critical role in aligning the visual features extracted by the InternViT model with the linguistic features processed by the Llama model. This alignment is essential for enabling seamless multimodal reasoning and generation tasks.

In the processing pipeline, text inputs are first tokenized and embedded using Llama 3's pre-trained tokenizer and embedding layer. These embedded tokens are then combined with the aligned visual features before being fed into the main Llama 3 model for multimodal understanding and generation.

\label{sec:architecture}

\section{Training Method}
\label{sec:method}
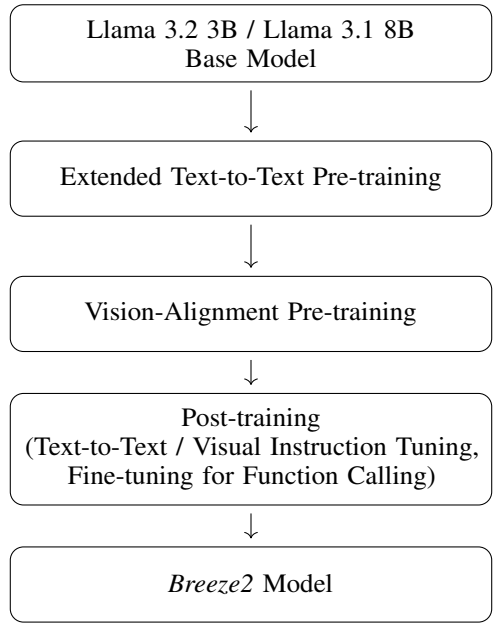
\begin{figure}[h]
    \centering
    \begin{tikzpicture}[
        node distance=1.8cm, 
        auto,
        block/.style={
            rectangle, 
            draw, 
            rounded corners=5pt, 
            text width=6cm, 
            align=center,
            minimum height=1.0cm,
            inner sep=0.2cm
        }
    ]
    \node[block] (A) {Llama 3.2 3B / Llama 3.1 8B\\Base Model};
    \node[block, below of=A] (B) {Extended Text-to-Text Pre-training};
    \node[block, below of=B] (C) {Vision-Alignment Pre-training};
    \node[block, below of=C] (D) {Post-training\\(Text-to-Text / Visual Instruction Tuning, Fine-tuning for Function Calling)};
    \node[block, below of=D] (E) {\textit{Breeze2} Model};
    
    \draw[->, shorten >=3pt, shorten <=3pt] (A) -- (B);
    \draw[->, shorten >=3pt, shorten <=3pt] (B) -- (C);
    \draw[->, shorten >=3pt, shorten <=3pt] (C) -- (D);
    \draw[->, shorten >=3pt, shorten <=3pt] (D) -- (E);
    \end{tikzpicture}
    \caption{Training Process of \textit{Breeze2} models. \textit{Breeze2} models were trained with a multi-stage training approach, starting from the Llama 3.2 3B and Llama 3.1 8B base models. The process starts with extended text-to-text pretraining to enhance Traditional Chinese language understanding. It is followed by vision-alignment pretraining to integrate visual features, and post-training for refining text and visual instruction tuning and function calling capabilities. }
    \label{fig:training_process}
\end{figure}

The training process for \textit{Breeze2} involves a comprehensive multi-stage approach, designed to enhance the model's capabilities in both textual and multimodal domains, with a specific focus on Traditional Chinese and Taiwan-centric content. As illustrated in \autoref{fig:training_process}, the training method consists of three main stages:

\begin{enumerate}
    \item \textbf{Extended Text-to-Text Pre-training:}
    \begin{itemize}
        \item Starting with either the Llama 3.2 3B or Llama 3.1 8B as the base model
        \item Extended pre-training to improve Traditional Chinese language understanding and generation
        \item Incorporation of Taiwan-specific knowledge and contexts
    \end{itemize}

    \item \textbf{Vision-Alignment Pre-training:}
    \begin{itemize}
        \item Integration of the InternViT-300M-448px vision encoder
        \item Aligning visual features with the language model through the randomly initialized vision-to-language projector
        \item Emphasis on multimodal understanding, particularly for Traditional Chinese text in images
    \end{itemize}

    \item \textbf{Post-training:}
    This final stage combines three critical components, with data from all components integrated for comprehensive training:
    \begin{itemize}
        \item \textbf{Text-to-Text Instruction Tuning:} Further refinement of the model's ability to follow text-based instructions and enhancing task-specific performance in Traditional Chinese
        \item \textbf{Visual Instruction Tuning:} Joint training on text and image inputs, improving the model's capability to generate coherent responses based on both textual and visual information, with a focus on Taiwan-related multimodal tasks
        \item \textbf{Fine-tuning for Function Calling:} Enhancing the model's ability to determine whether and which functions to call in response to Traditional Chinese language prompts
    \end{itemize}
\end{enumerate}

This comprehensive training approach ensures that \textit{Breeze2} not only excels in Traditional Chinese language processing but also demonstrates strong performance in multimodal tasks and function calling, particularly those relevant to Taiwan-specific contexts.

\section{Pre-training}
\label{sec:pretraining}
Building upon the architectural foundation described in the previous section, the \textit{Breeze2} series leverages the powerful Llama 3 series models as its core. This chapter delves deeper into the sophisticated pre-training process that not only elevates \textit{Breeze2} from a basic language model to a versatile multimodal system but also significantly enhances its capabilities in Traditional Chinese language processing and understanding of Taiwan-specific contexts.

The extended pre-training strategy of the \textit{Breeze2} series is meticulously designed to strengthen the model's proficiency in Traditional Chinese, with a particular focus on Taiwan-related knowledge. This process consists of two distinct stages:

\begin{enumerate}
    \item \textbf{Text-to-Text Training:} This initial stage focuses on enhancing the language understanding and generation capabilities inherited from the Llama 3 base model. Special emphasis is placed on improving the model's grasp of Traditional Chinese linguistic nuances and Taiwan-specific terminology and concepts.
    
    \item \textbf{Vision Alignment:} The second stage involves a crucial pre-training step that bridges the text-to-text base model with the InternViT image encoder, as introduced earlier. This stage is key to enabling \textit{Breeze2}'s multimodal functionalities, with particular attention to visual understanding in a Traditional Chinese and Taiwanese cultural context.
\end{enumerate}

Furthermore, this chapter will provide a comprehensive overview of the carefully curated datasets employed in each stage of pre-training. These datasets are specifically selected to include a rich variety of Traditional Chinese texts and images relevant to Taiwan's cultural, social, and historical landscape. We will also detail the training methods and techniques utilized to optimize the model's performance across both textual and visual domains, with a consistent focus on Traditional Chinese and Taiwan-centric content.

\subsection{Text-to-text Pre-training Data for Localization}
\label{sec:pretrain-data-loc}

For the text-to-text pre-training stage focused on Traditional Chinese and Taiwan-specific content, we curated a diverse dataset from various sources. \autoref{tab:pretrain-data} summarizes the composition of this dataset.

\begin{table}[h]
\centering
\caption{Composition of text-to-text pretraining dataset}
\label{tab:pretrain-data}
\begin{tabular}{|l|r|l|}
\hline
\textbf{Category} & \textbf{Size (GB)} & \textbf{Description} \\
 \hline \hline
Web Crawl & 421.3 & CommonCrawl  (CC100 ZHTW \citep{conneau-etal-2020-unsupervised}, \\ 
 & & C4 ZHTW \citep{dodge-etal-2021-documenting}, FineWeb-zhtw \citep{lin2024finewebzhtwscalablecurationtraditional}) \\ 
 & & and other miscellaneous web data \\  \hline
Academic & 94.0 & Theses, research plans, and reports \\ \hline
Code & 110.0 & Programming-related content (The Stack \citep{Kocetkov2022TheStack}) \\ \hline
News & 131.0 & News articles \\ \hline
Dialogue & 37.0 & Transcribed conversation, legislative records \\ \hline
Blog & 20.7 & Blog content \\ \hline
Forum & 60.0 & Q\&A content, BBS, forum discussions \\ \hline
Judicial & 21.4 & Legal judgments and documents \\ \hline
Books & 1.6 & Book excerpts and summaries \\ \hline
Wiki & 1.1 & Wikipedia content in Traditional Chinese \\  \hline 
\hline
\textbf{Total} & \textbf{898.1} & \\
\hline
\end{tabular}
\end{table}

This carefully curated dataset encompasses a wide range of content types, ensuring comprehensive coverage of Traditional Chinese language usage and Taiwan-specific knowledge. The dataset includes web crawl data, academic publications, code repositories, news articles, dialogues, social media content, forum discussions, legal documents, and literary works. This diverse collection aims to enhance the model's understanding of various domains and linguistic styles prevalent in Taiwan's cultural and social context.

\subsection{Training Details of Text-to-text Pre-training}

The text-to-text pre-training stage for \textit{Breeze2} was conducted using 5,760 H100-GPU hours for the 3B version and 16,620 H100-GPU hours for the 8B version. This training process allowed for comprehensive learning of the model's language understanding and generation capabilities, particularly for Traditional Chinese and Taiwan-specific contexts.

Key training parameters and configurations include:

\begin{itemize}
    \item \textbf{Parallelism Strategy:} For the 3B version, only data parallelism was employed during training. In contrast, the 8B version required a tensor parallelism size of 2, in addition to data parallelism, to efficiently manage the larger model size.
    
    \item \textbf{Precision and Optimization:} The model utilized bfloat16 precision and Transformer Engine's FP8 training support, as provided by NVIDIA's NeMo Framework \citep{kuchaiev2019nemotoolkitbuildingai}.
    
    \item \textbf{Batch Size and Sequence Length:} A global batch size of 1,024 was employed, with a maximum sequence length of 8,192 tokens. This configuration allowed for maintaining a balance between computational efficiency and model performance.
    
    \item \textbf{Learning Rate and Scheduler:} The training utilized a distributed fused Adam optimizer with an initial learning rate of $1\times10^{-5}$. A cosine annealing learning rate scheduler was implemented, featuring a warmup period of 500 steps and a minimum learning rate of $1\times10^{-6}$.
\end{itemize}

These training details provide insight into the pre-training process of \textit{Breeze2}, leveraging the capabilities of NVIDIA's NeMo Framework. The difference in training hours and parallelism strategies between the 3B and 8B versions reflects the adaptations necessary for training larger language models efficiently.

\subsection{Pre-training Data for Visual Alignment}
\label{sec:pretrain-data-vis}



To achieve effective visual alignment, we implemented a comprehensive 2-phase pre-training approach. This method ensures that our models are well-prepared for a variety of visual tasks.

In the first phase, we focused on training only the MLP layers. This initial step allows the model to establish a foundational understanding of the data, utilizing approximately 3.42M data points. Following this, we transitioned to the second phase, which involves full weights pre-training. During this phase, all components of the model, including ViT, MLP, and LLM, are trainable, utilizing approximately 10.89M data points. This holistic training approach ensures that the model can leverage the strengths of each component to achieve superior performance.

The pre-training dataset for our \textit{Breeze2} model is sourced from a diverse array of publicly accessible datasets. These datasets cover a wide range of tasks, such as captioning, visual question answering, detection, grounding, and OCR (Optical Character Recognition). In tables \autoref{tab:Phase1 pretrain-datasets} and \autoref{tab:Phase2 pretrain-datasets}, we detail the specific datasets used in each phase of our pre-training process.


\begin{table}[h]
\centering
\caption{Datasets for pretraining the vision-to-language projector}
\label{tab:Phase1 pretrain-datasets}
\begin{tabular}{|l|l|l|}
\hline
\textbf{Category} & \textbf{\# of Data} & \textbf{Datasets} \\ \hline  \hline
 Captioning& 2.8M & 
\begin{tabular}[c]{@{}l@{}}
ALLVA \citep{chen2024allava}, ShareGPT4V \citep{chen2023sharegpt4vimprovinglargemultimodal},\\
ShareCaptioner \citep{chen2023sharegpt4vimprovinglargemultimodal},\\
Traditional Chinese version of LAION \citep{schuhmann2022laion5bopenlargescaledataset},\\
Traditional Chinese version of COYO \citep{kakaobrain2022coyo-700m},\\
Traditional Chinese version of GRIT-Caption \citep{peng2023kosmos2groundingmultimodallarge}\\ 
\end{tabular} \\  \hline
General QA  & 0.22M & VQAv2 \citep{balanced_vqa_v2}\\  \hline
OCR  & 0.4M &  Traditional Chinese version of SynthDoG \citep{kim2022ocrfreedocumentunderstandingtransformer} \\
\hline  \hline
  \textbf{Total}  & \textbf{3.42M} & \\ \hline
\end{tabular}
\end{table}

\begin{table}[h]
\centering
\caption{Datasets used for end-to-end ViT, MLP, LLM alignment}
\label{tab:Phase2 pretrain-datasets}
\begin{tabular}{|l|l|l|}
\hline
\textbf{Category} & \textbf{\# of Data} & \textbf{Datasets} \\ \hline \hline
 Captioning& 3.42M & 
\begin{tabular}[c]{@{}l@{}}
ShareGPT4V,
ShareCaptioner, ALLVA,\\
ShareGPT4o \citep{chen2024fargpt4vclosinggap}, \\
TaiTravel \citep{TaiTravel}, \\
Datasets from the cauldron \citep{laurençon2024matters}, \\
Traditional Chinese version of ShareGPT4V\\
\end{tabular} \\  \hline
General QA  & 1.02M & VQAv2, Datasets from the cauldron\\  \hline
Detection & 0.2M & AS-V2 \citep{wang2024allseeing_v2}\\ \hline
OCR  & 2.05M& 
\begin{tabular}[c]{@{}l@{}}ArT \citep{chng2019icdar2019robustreadingchallenge}, COCO-Text \citep{veit2016cocotextdatasetbenchmarktext}, \\
Datasets from the cauldron, In-house OCR dataset, \\
SynthDoG, Traditional Chinese version of SynthDoG\\
\end{tabular} \\ \hline
Reasoning & 0.39M & Datasets from the cauldron\\ \hline
Science & 0.02M & Datasets from the cauldron\\ \hline
Chart & 3.17M & MMC-Inst \citep{liu2024mmcadvancingmultimodalchart}, Datasets from the cauldron\\ \hline
Knowledge & 0.52M &
\begin{tabular}[c]{@{}l@{}}WIT \citep{10.1145/3404835.3463257},\\  In-house Taiwan-knowledge caption dataset\\
\end{tabular} \\  
\hline \hline
  \textbf{Total}  & \textbf{10.89M} & \\ 
 
  \hline
\end{tabular}
\end{table}

By leveraging these diverse datasets, our pre-training process equips the \textit{Breeze2} model with the ability to handle a wide range of visual tasks, ensuring robust performance across different applications.

\subsection{Training Details of Visual Alignment Pre-training}
\label{sec:pretrain-hyp}

The visual alignment pre-training process consists of two sequential phases to enhance training efficiency and performance based on our experimental results:

\begin{enumerate}
\item \textbf{Phase 1: MLP Phase}: The MLP projector was trained from scratch for one epoch. The AdamW optimizer was employed with a base learning rate of $1 \times 10^{-5}$. The learning rate schedule consisted of a 10\% linear warm-up period, followed by a constant learning rate. This phase utilized a batch size of 1024.
\item \textbf{Phase 2: Full-weight Phase}: Subsequent to the MLP phase, full-weight training was conducted for 2 epochs. The AdamW optimizer was again utilized, this time with a base learning rate of $1 \times 10^{-6}$. The learning rate schedule mirrored that of the MLP phase, incorporating a 10\% linear warm-up followed by a constant rate. The batch size of 1024 was maintained for this phase as well.
\end{enumerate}

\section{Post-training}

\textit{Breeze2} employs a tuning-based methodology to achieve both instruction-following and function-calling capabilities. This approach involves fine-tuning pre-trained base models using specifically designed prompt templates, as detailed in Section \ref{sec:prompt-template}. We have curated three distinct post-training datasets: one for instruction following (Section \ref{sec:post-data-if}), one for visual instruction following (Section \ref{sec:post-data-vis}), and one for function calling (Section \ref{sec:post-data-fc}). We applied supervised fine-tuning (SFT) to the base model using a combination of these three post-training datasets. The details of the post-training are shown below (Section \ref{sec:post-hyp})

\subsection{Prompt Template Design}
\label{sec:prompt-template}

For consistency, we primarily adopted the prompt template from Llama 3, which is based on the Chat Markup Language (ChatML), a widely adopted format introduced by OpenAI. We then made slight adjustments to fit the needs for enabling vision-aware and function-calling capabilities. The key differences are as follows:

\begin{itemize}
\item \textbf{Image placeholder.} We introduce the special tokens \texttt{<|start\_img|>}, \texttt{<|img|>}, and \texttt{<|end\_img|>}. A sequence of \texttt{<|img|>} tokens represents the patches of a provided image, while the \texttt{<|start\_img|>} and \texttt{<|end\_img|>} tokens enclose this sequence to indicate the position of the image placeholder.

\item \textbf{Bounding box placeholder.} We introduce the special tokens \texttt{<|start\_bbox|>} and \texttt{<|end\_bbox|>}. These tokens enclose the list of coordinates of the bounding box. The coordinates are represented by four values: the x-coordinate of the top-left point, the y-coordinate of the top-left point, the x-coordinate of the bottom-right point, and the y-coordinate of the bottom-right point. All coordinates are normalized between 0 and 1000 and represented as integers.

\item \textbf{Providing function descriptions.} Based on the investigation in \citep{chen2024enhancing}, functions provided in a dedicated role and the system role exhibit similar capabilities in terms of function-calling accuracy, but relevance detection is superior when functions are provided in the dedicated role. For consistency with Llama 3, we finally chose to provide function descriptions in the system role and compromised on relevance detection by introducing the Decision Token proposed by \citep{chen2024enhancing}. To provide the function descriptions for enabling function calling, we prepend the list of the function description following the ``Customized Functions:''.

\item \textbf{Decision Token.} We introduce the special tokens \texttt{<|use\_tool|>} and \texttt{<|answer|>} as the Decision Token \citep{chen2024enhancing}. The Decision Token is the first generated token when functions are given, determining the action of models. If the model chooses to provide a direct answer, it outputs \texttt{<|answer|>} first; if it chooses function calling, it outputs \texttt{<|use\_tool|>} first. The Decision Token also facilitates the creation of non-function-call data from function-called data, significantly enhancing relevance detection (see Section \ref{sec:post-data-fc} for details). 

\item \textbf{Supporting function calls and responses in parallel.} Llama 3 supports only a single function call and response. We extend this to support function calls and responses in parallel. The list of function calls is provided following the special token \texttt{<|python\_tag|>}. The list of function responses is put in the \texttt{ipython} role, with the order item by item corresponding to function calls.
\end{itemize}

More details are provided in the GitHub repository\footnote{\url{https://github.com/mtkresearch/mtkresearch/blob/main/doc/format\_prompt\_v3.md}}. To facilitate developers' access to \textit{Breeze2}, we also provide a Python package\footnote{\url{https://github.com/mtkresearch/mtkresearch}} to handle the complex conversion from conversation turns to the conditional prompt given to the model.

\subsection{Post-training Data for Instruction Following}
\label{sec:post-data-if}

We employ SlimOrca \citep{SlimOrca, mukherjee2023orca, longpre2023flan} for post-training fine-tuning. SlimOrca, a refined subset of OpenOrca, incorporates an additional filtering mechanism to exclude potentially mislabeled data points. Similar to Orca, SlimOrca comprises an augmented FLAN dataset that emphasizes solving questions through reasoning chains. The diverse system prompts within the corpus enhance the model's attentiveness to descriptions. Our training utilizes the entire dataset, which includes 500k data points in English. Preliminary experiments indicate that the fine-tuned model, extensively trained on English instances, demonstrates strong performance in Traditional Chinese. We attribute this performance to the model's knowledge transfer ability, which is enhanced by intensive pre-training on Traditional Chinese.

\subsection{Post-training Data for Visual Instruction Following}
\label{sec:post-data-vis}

To enhance the model's ability to follow visual instructions, we utilized a comprehensive set of datasets for post-training. This includes all datasets from the 2-phase pre-training, each with a selected portion to ensure a balanced and diverse training set.

In addition to the pre-training datasets, we incorporated several high-quality chat datasets to further improve the model's chat capabilities. These additional datasets include:

\begin{itemize}
    \item LLaVA \citep{liu2023llava}
    \item LLaVAR-2 \citep{zhou2024highqualitytextrichimageinstruction}
    \item High-quality chat datasets from Canbrain-1 \citep{tong2024cambrian1fullyopenvisioncentric}
\end{itemize}

To specifically enhance the model's ability to handle image-based Chinese chat, we also included an in-house Taiwan-knowledge chat dataset. This comprehensive post-training dataset comprises a total of 6M data points, ensuring that the model is well-equipped to handle a wide range of visual instruction tasks.

By leveraging these diverse and high-quality datasets, our post-training process significantly boosts the model's performance in visual instruction following, making it more robust and versatile in real-world applications. 


\subsection{Post-training Data for Function Calling}
\label{sec:post-data-fc}

To enable function calling capabilities, we curated high-quality function-calling datasets, including:

\begin{itemize}
\item \texttt{FC-110k}: This dataset comprises 110k high-quality instances, filtered from a combination of APIGen \cite{liu2024apigen} and the glaive-function-calling-v2 dataset\footnote{\url{https://huggingface.co/datasets/glaiveai/glaive-function-calling-v2}}.

\item \texttt{FC-TC-19k}: We generated 19k function-calling instances in Traditional Chinese using synthetic translation methods proposed by \citet{chen2024enhancing}. Instances were selected from \texttt{FC-110k} while striving to maintain a balanced distribution of function types among the called functions.

\item \texttt{FC-NF-10k}: Using synthetic methods proposed by \citet{chen2024enhancing}, we generated 10k instances of non-function-call data from \texttt{FC-110k} and \texttt{FC-TC-19k}. The ratio between English and Traditional Chinese instances is 9:1. Specifically, to generate non-function-call data, consider an example with three functions: \texttt{func\_A}, \texttt{func\_B}, and \texttt{func\_C}. If \texttt{func\_A} is helpful and called in the original data, we create non-function-call data by removing \texttt{func\_A} and only using \texttt{func\_B} and \texttt{func\_C}. This ensures that function calling is not triggered.
\end{itemize}

\subsection{Training Details of Post-training}
\label{sec:post-hyp}

We employed the AdamW optimizer with a learning rate of \(1 \times 10^{-6}\). The learning rate was warmed up during the first 10\% of the training steps. Gradient clipping was applied with a maximum norm of 1. The model was trained for a total of 4 epochs with a batch size of 1024.

\section{Benchmarks and Results}
\label{sec:benchmarks}

\textit{Breeze2} underwent comprehensive evaluation across numerous established benchmarks. These assessments cover a wide spectrum of tasks, including genera knowledge, long context capabilities, function calling, and visual comprehension.

\subsection{Taiwan General Knowledge and Instruction-Following Benchmarks}
\label{sec:benchmark-tw} 

For general knowledge of Taiwan, we tested on TMMLU+ \citep{ikala2023eval, model7c2023}. The dataset composed of 67 subjects covering disciplines from STEM to humanities, and includes professional domain knowledge as well. As expected, the \textit{Breeze2} models significantly outperformed comparable models. Additionally, to serve as a baseline for our evaluation, we measured the chatting ability using MT-Bench-tw, which is a translated version of MT-Bench \citep{zheng2024judging} with detailed manual rewriting. MT-Bench-tw along with MT-Bench highlights multi-turn dialogue capabilities of the language model. Again as expected, given that \textit{Breeze2} are uniformly optimized for knowledge, function calling, and vision as opposed to chat, its performance in the chat arena was only decent. Despite this, perhaps due to its superior Traditional Chinese knowledge, our 8B model still beat Llama-3.1 8B in MT-Bench-tw.

\autoref{tab:general-results} demonstrates the evaluation results.

\begin{table}[h!]
\centering
\caption{Comparison among comparable models across generic and Taiwan-specific language knowledge benchmarks and chat-specific benchmarks. The judge model for MT-Bench-tw and MT-Bench is GPT-4o.}
\begin{tabular}{c|c|c|c|c}
\hline
 & TMMLU+ (0-shot) & MMLU (0-shot) & MT-Bench-tw & MT-Bench \\ \hline
Llama-3.2 3B & 35.3 & 53.3 & \textbf{4.49} & \textbf{6.99} \\ 
Breeze2 3B & \textbf{42.7} & \textbf{59.9} & 4.26 & 5.96 \\ 
\hline
Breeze 7B & 42.7 &61.7 & 5.43  &6.49 \\ 
Llama-3.1 8B & 39.2 & 59.8 & 5.54 & \textbf{7.21} \\
Breeze2 8B & \textbf{46.6} & \textbf{67.0} & \textbf{5.79} & 7.05 \\ 
\hline \hline
GPT-4o-mini & 38.26 & 42.62 & 7.86 & 8.24 \\
\hline
\end{tabular}

\label{tab:general-results}
\end{table}

\subsection{Long Context Benchmarks}
\label{sec:benchmark-long}

We use the synthetic passkey retrieval task to evaluate long context capabilities of the model. The passkey retrieval task is designed to assess the model's ability to accurately retrieve and utilize a passkey or specific information embedded within lengthy irrelevant text. This task is particularly relevant in scenarios where the model needs to process and understand large volumes of text to extract critical information.

To thoroughly evaluate the model's performance across different parts of the 128k context, we divide the context into 16 bins, each covering an 8k token range. Bin 0 represents the first 8k tokens (0-8k), while Bin 15 represents the last 8k tokens (120k-128k). The passkey is placed at different positions within these bins to test the model's retrieval capability throughout the entire context length. The results are listed in \autoref{tab:passkey-results-combined}.

Evaluation results show the model's awareness of information within the long context at different relative positions throughout the 128k range.

\begin{table}[h!]
\centering
\caption{Comparison of \textit{Breeze2} 3B and 8B models on passkey retrieval task across all 16 bins (0-128k tokens): Both models show strong performance, with 8B maintaining perfect accuracy in later bins.}
\begin{tabular}{c|c|c|c|c|c|c|c|c}
\hline
\multirow{2}{*}{Model} & \multicolumn{8}{c}{Retrieval accuracy-128k (\%)} \\ \cline{2-9}
 & \stackunder[1pt]{Bin 0}{\scriptsize(0-8k)} & \stackunder[1pt]{Bin 1}{\scriptsize(8-16k)} & \stackunder[1pt]{Bin 2}{\scriptsize(16-24k)} & \stackunder[1pt]{Bin 3}{\scriptsize(24-32k)} & \stackunder[1pt]{Bin 4}{\scriptsize(32-40k)} & \stackunder[1pt]{Bin 5}{\scriptsize(40-48k)} & \stackunder[1pt]{Bin 6}{\scriptsize(48-56k)} & \stackunder[1pt]{Bin 7}{\scriptsize(56-64k)} \\ \hline
Breeze2 3B & 100 & 100& 100& 100& 100& 100& 95& 95\\ 
Breeze2 8B & 100& 100& 95& 100& 95 &100 & 100&100 \\ \hline
\multicolumn{9}{c}{} \\ \hline
\multirow{2}{*}{Model} & \multicolumn{8}{c}{Retrieval accuracy-128k (\%)} \\ \cline{2-9}
 & \stackunder[1pt]{Bin 8}{\scriptsize(64-72k)} & \stackunder[1pt]{Bin 9}{\scriptsize(72-80k)} & \stackunder[1pt]{Bin 10}{\scriptsize(80-88k)} & \stackunder[1pt]{Bin 11}{\scriptsize(88-96k)} & \stackunder[1pt]{Bin 12}{\scriptsize(96-104k)} & \stackunder[1pt]{Bin 13}{\scriptsize(104-112k)} & \stackunder[1pt]{Bin 14}{\scriptsize(112-120k)} & \stackunder[1pt]{Bin 15}{\scriptsize(120-128k)} \\ \hline
Breeze2 3B & 95& 95& 100&100& 100& 100& 95&90 \\ 
Breeze2 8B & 100&100&100&100&100&100&100&100 \\ \hline
\end{tabular}
\label{tab:passkey-results-combined}
\end{table}

\subsection{Function Calling Benchmarks}
\label{sec:benchmark-fc}

To evaluate the performance of function calling, we utilized the Berkeley Function Calling Leaderboard (BFCL) \citep{berkeley-function-calling-leaderboard} for English contexts. Additionally, we employed the Function Calling Leaderboard for ZHTW \citep{fc-leaderboard-zhtw}, a translated version of the BFCL, to assess performance in Traditional Chinese contexts. The evaluation metrics are described below.

\textbf{AST Accuracy (\%)}: This metric evaluates the structural correctness of language model outputs for function-calling tasks by comparing the Abstract Syntax Tree (AST) representations of generated and target function calls. It includes four problem types—Simple Function, Multiple Function, Parallel Function, and Parallel Multiple Function—categorized based on the combination of the number of provided functions and function calls.

\textbf{Executable Accuracy (\%)}: This metric assesses performance by executing the generated APIs and comparing the results with the expected outcomes. It also includes the same four problem types as AST Accuracy.

\textbf{Relevance Detection (\%)}: This metric measures the success rate of not generating any function call when none of the provided functions are relevant. This scenario helps determine whether a model will hallucinate its functions and parameters when the provided functions are irrelevant to the user's query.

\autoref{tab:benchmark-fc-en} and \autoref{tab:benchmark-fc-tc} present the evaluation results, where our \textit{Breeze2} 8B is comparable to the commercial model GPT-4o-mini on both benchmarks in English and Traditional Chinese.

\begin{table}[h!]
\centering
\caption{Results of BFCL in English contexts. S., M., P., and P.M. represent Simple Function, Multiple Functions, Parallel Functions, and Parallel Multiple Functions, respectively. The version of GPT-4o-mini is 2024-07-18.}
\begin{tabular}{c|c|c|c|c|c|c|c|c|c|c}
\hline
~ & \textbf{Overall} & \multicolumn{4}{c|}{\textbf{AST Accuracy}} & \multicolumn{4}{c|}{\textbf{Executable Accuracy}} & \textbf{Relevance} \\ 
~ & \textbf{Accuracy} & S. & M. & P. & P.M. & S. & M. & P. & P.M. & \textbf{Detection} \\ \hline
Breeze2 3B  & 85 & 92 & 92 & 84 & 81 & 95 & 92 & 84 & 78 & 65 \\
\hline
Llama-3.1 8B  & 57 & 56 & 54 & 49 & 35 & 59 & 58 & 54 & 30 & 94 \\
Gorilla-v2 7B & 86 & 94 & 96 & 87 & 86 & 97 & 96 & 80 & 75 & 60 \\
Breeze2 8B  & \textbf{87} & 91 & 93 & 86 & 81 & 96 & 92 & 82 & 65 & 83 \\
\hline \hline
GPT-4o-mini  & 87 & 91 & 93 & 90 & 84 & 97 & 90 & 86 & 65 & 77 \\ 
\hline
\end{tabular}
\label{tab:benchmark-fc-en}
\end{table}

\begin{table}[h!]
\centering
\caption{Results of the Function Calling
Leaderboard for ZHTW in Traditional Chinese contexts. S., M., P., and P.M. represent Simple Function, Multiple Functions, Parallel Functions, and Parallel Multiple Functions, respectively. The version of GPT-4o-mini is 2024-07-18.}
\begin{tabular}{c|c|c|c|c|c|c|c|c|c|c}
\hline
~ & \textbf{Overall} & \multicolumn{4}{c|}{\textbf{AST Accuracy}} & \multicolumn{4}{c|}{\textbf{Executable Accuracy}} & \textbf{Relevance} \\ 
~ & \textbf{Accuracy} & S. & M. & P. & P.M. & S. & M. & P. & P.M. & \textbf{Detection} \\ \hline
Breeze2 3B  & 76 & 82 & 83 & 79 & 66 & 89 & 82 & 76 & 65 & 65 \\ 
\hline
Gorilla-v2 7B & 76 & 85 & 87 & 73 & 68 & 92 & 92 & 62 & 73 & 54 \\ 
Breeze2 8B  & \textbf{80} & 84 & 88 & 86 & 67 & 89 & 82 & 78 & 63 & 76 \\ \hline \hline
GPT-4o-mini  & 81 & 86 & 89 & 86 & 78 & 88 & 86 & 70 & 58 & 67 \\
\hline
\end{tabular}
\label{tab:benchmark-fc-tc}
\end{table}

\subsection{Vision Understanding Benchmarks}
\label{sec:benchmark-vis}
To evaluate the vision capability of our models, we have tested them on MMMU \citep{yue2023mmmu} and our proposed TMMBench\footnote{\url{https://github.com/mtkresearch/TMMBench}}. MMMU is a vision-language benchmark that contains questions from college-level exams and textbooks. It includes diverse image types and focuses on complex reasoning tasks that require specialized knowledge. To evaluate the comprehension of traditional Chinese and Taiwan-specific knowledge, we have developed an additional benchmark named TMMBench. TMMBench consists of traditional Chinese questions covering topics related to Taiwan, such as Taiwanese attractions, daily life, tables or diagrams containing traditional Chinese, and questions from Taiwan's university entrance exams. We plan to release TMMBench in the near future. 
The evaluation results are shown in \autoref{tab:vl-results}, where our \textit{Breeze2} 8B outperforms the larger Llama-3.2 11B model across both benchmarks.

\begin{table}[h!]
\centering
\caption{Comparison among comparable models across various vision-language benchmarks}
\begin{tabular}{c|c|c}
\hline
 & MMMU & TMMBench \\ \hline
Breeze2  3B &38.2 &41.7 \\
\hline
Llama-3.2 11B & 41.7 & 42.4 \\
Breeze2 8B & \textbf{43.2} & \textbf{45.2}  \\ \hline \hline
GPT-4o-mini & 59.4 & 55.5 \\ 
\hline
\end{tabular}

\label{tab:vl-results}
\end{table}

\section{Demonstration}

\subsection{Case Study: Text-to-text Instruction Following}
\begin{CJK*}{UTF8}{bsmi}
To demonstrate the enhanced capabilities of \textit{Breeze2} 3B in handling Taiwan-specific contexts and following complex instructions, we conducted two comparative case studies. The first focuses on knowledge about Taiwanese night markets, while the second examines the model's ability to generate more formal and polite text based on given instructions.

\subsubsection{Case Study 1: Taiwanese Night Markets}

We presented both models with the following query in Traditional Chinese:

\begin{quote}
請改寫以下短文，使其具備關於台灣夜市的具體細節，例如著名的夜市名稱。

台灣的夜市文化是台灣人生活中不可或缺的一部分。夜市不僅提供各種美食，還有各式各樣的商品和娛樂活動。每當夜幕降臨，夜市便開始熱鬧起來，吸引了大量的遊客和當地居民。夜市裡的攤販們熱情地招呼顧客，讓人感受到濃厚的人情味。夜市裡的美食總能滿足不同人的口味。夜市不僅是購物和品嚐美食的好地方，也是體驗台灣文化的絕佳場所。
\end{quote}

\begin{quote}
    \textit{English translation:\newline Please rewrite the following short passage to include specific details about Taiwanese night markets, such as the names of famous night markets.\newline Night market culture is an integral part of Taiwanese life. Night markets not only offer a variety of delicious food but also various goods and entertainment activities. As night falls, these markets come alive, attracting numerous tourists and locals alike. The vendors in the night markets enthusiastically greet customers, creating a warm and friendly atmosphere. The food at night markets can satisfy different tastes. Night markets are not only great places for shopping and tasting delicious food but also excellent venues for experiencing Taiwanese culture.}
\end{quote}

The responses from both models were then analyzed for accuracy and richness of Taiwan-specific information. Below are the original responses from both models:

\begin{quote}
\textbf{Breeze2 3B:} 台灣的夜市文化是台灣人生活中不可或缺的一部分。著名的夜市如士林夜市、饒河街夜市和羅東夜市等，每當夜幕降臨，夜市便開始熱鬧起來，吸引了大量的遊客和當地居民。夜市裡的攤販們熱情地招呼顧客，讓人感受到濃厚的人情味。夜市裡的美食總能滿足不同人的口味。夜市不僅是購物和品嚐美食的好地方，也是體驗台灣文化的絕佳場所。

\end{quote}

\begin{quote}
\textbf{Llama 3.2 3B Instruct:} 台灣的夜市文化是台灣人生活中不可或缺的一部分。夜市不僅提供各種美食，還有各式各樣的商品和娛樂活動。每當夜幕降臨，夜市便開始熱鬧起來，吸引了大量的遊客和當地居民。夜市裡的攤販們熱情地招呼顧客，讓人感受到濃厚的人情味。如，台北的夜市如士林夜市、電信夜市、世貿夜市等，都是人氣的熱門夜市之一。夜市裡的美食總能滿足不同人的口味，從海鮮到甜點，還有各種特殊的夜市特產。夜市不僅是購物和品嚐美食的好地方，也是體驗台灣文化的絕佳場所。
\end{quote}

The responses reveal a significant difference in the depth and accuracy of knowledge about Taiwanese night markets between the two models. \textit{Breeze2} 3B demonstrated a more comprehensive and accurate understanding of the topic. It correctly named well-known night markets across Taiwan, including Shilin Night Market (士林夜市), Raohe Street Tourist Night Market (饒河街夜市), and Luotong Night Market (羅東夜市). These night markets span multiple cities and regions in Taiwan, showcasing \textit{Breeze2} 3B's broad understanding of the country's night market culture.

In contrast, Llama 3.2 3B Instruct only correctly identified one night market, Shilin Night Market (士林夜市). Moreover, it generated two non-existent night markets, ``Telecom Night Market'' (電信夜市) and ``World Trade Night Market'' (世貿夜市), which do not exist in reality.

\subsubsection{Case Study 2: Formalizing Email Content}

For this case study, we presented both models with the following task in Traditional Chinese:

\begin{quote}
使以下郵件的風格更加正式，並使用敬語。

嗨，John，

我想跟你確認一下我們下週的會議時間。你方便的話，能不能告訴我你什麼時候有空？另外，我們需要討論一下新項目的進度。

謝謝， Tom
\end{quote}

\begin{quote}
    \textit{English translation: \newline Make the following email more formal and use honorific language.\newline Hi John,\newline I want to confirm our meeting time for next week. If it's convenient for you, could you tell me when you're available? Also, we need to discuss the progress of the new project.\newline Thanks, Tom}
\end{quote}

The responses from both models were then analyzed for formality, politeness, and overall improvement in language. Below are the original responses from both models:

\begin{quote}
\textbf{Breeze2 3B:}

您好，John，

我想與您確認一下我們下週的會議時間。您方便的話，能否告訴我您什麼時候有空？另外，我們需要討論一下新項目的進度。

謝謝，Tom




\end{quote}

\begin{quote}
\textbf{Llama 3.2 3B Instruct:}

嗨，John，

我想跟您確認一下我們下週的會議時間。您是否有空的話，能否告訴我您什麼時候有空？

另外，我們還需要討論一下新項目的進度。望您能夠提供相關的資訊。

謝謝，Tom
\end{quote}

The responses reveal a notable difference in the quality of formalization and use of honorific language between the two models. \textit{Breeze2} 3B demonstrated a more natural and appropriate transformation of the original text:

\begin{enumerate}
    \item It changed the greeting from ``嗨'' (Hi) to ``您好'', which is more formal in Traditional Chinese business correspondence.
    \item It consistently used the honorific ``您'' instead of ``你'' throughout the email.
    \item The sentence structure was improved for clarity and formality without unnecessary additions.
\end{enumerate}

In contrast, Llama 3.2 3B Instruct made some improvements but also introduced issues:

\begin{enumerate}
    \item It retained the informal greeting``嗨'' (Hi), which is inconsistent with the request for a more formal tone.
    \item While it did use the honorific ``您'', it created a redundant and awkward sentence: ``您是否有空的話，能否告訴我您什麼時候有空？'' (If you are available, could you tell me when you are available?), which is both repetitive and illogical.
    \item It added an unnecessary sentence at the end: ``望您能夠提供相關的資訊。'' (Hope you can provide relevant information.), which wasn't part of the original message and doesn't significantly enhance formality.
\end{enumerate}

\subsubsection{Analysis and Implications}
These case studies underscore the effectiveness of \textit{Breeze2} 3B's specialized pre-training in capturing nuanced, location-specific knowledge and handling complex language tasks in Traditional Chinese. In the first case study, the model demonstrated a comprehensive understanding of Taiwanese night markets, accurately naming multiple locations across different cities. This showcases its enhanced capability in handling queries that require deep understanding of local cultural contexts.

In the second case study, \textit{Breeze2} 3B exhibited superior ability in refining language for formal communication. The model demonstrated a better understanding of formal language conventions and was able to make appropriate adjustments without altering the core message or introducing redundancies.

These case studies highlight \textit{Breeze2} 3B's advanced capabilities in handling Taiwan-specific contexts and complex language tasks in Traditional Chinese. The model's ability to provide accurate, contextually appropriate responses demonstrates its potential for enhancing a wide range of applications that require nuanced understanding of local culture and language use.
\end{CJK*}

\subsection{Case Study: Visual Instruction Following}
\begin{CJK*}{UTF8}{bsmi}

To demonstrate the visual instruction-following capability of Breeze2 8B, we provide an example that showcases its ability to understand and process visual information in conjunction with Traditional Chinese text queries.

\begin{figure}[!ht]
    \centering
    \includegraphics[width=0.6\textwidth]{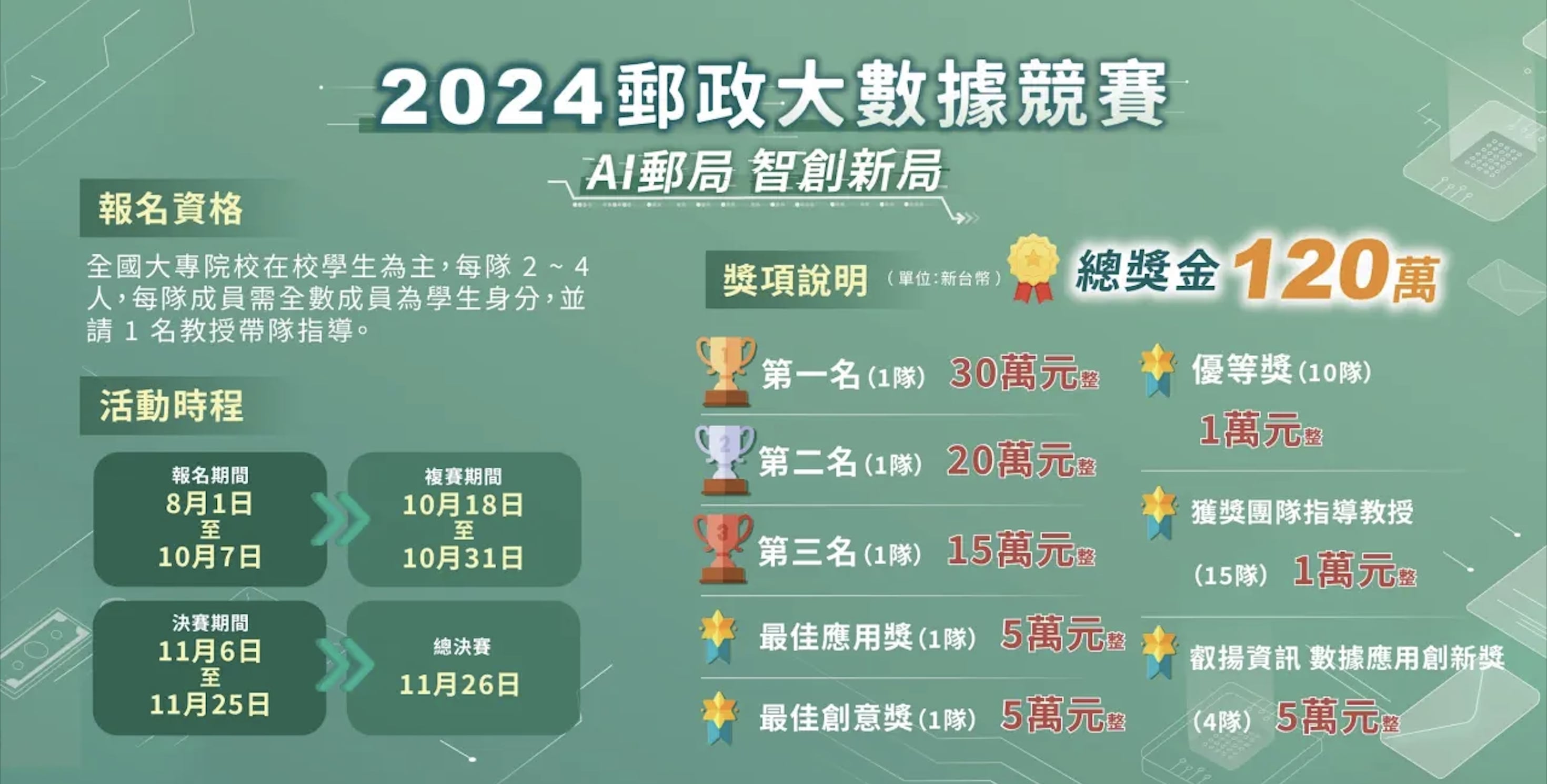} 
    \caption{Image for demo showing a prize distribution table.}
    \label{fig:img_example}
\end{figure}

We presented the model with the following visual query in Traditional Chinese using \autoref{fig:img_example}:

\begin{quote}
<image>\newline
請問前三名總共可獲得多少錢？
\end{quote}

\begin{quote}
    \textit{English translation:\newline<image>\newline How much money will the top three get in total?}
\end{quote}

The response from Breeze2 8B demonstrates its capability to accurately interpret the question and extract relevant information from the image:

\begin{quote}
\textbf{Breeze2 8B:}\newline
第一名可獲得30萬元，第二名可獲得20萬元，第三名可獲得15萬元。前三名總共可獲得65萬元。
\end{quote}

\begin{quote}
    \textit{English translation:\newline
    The first place winner gets 300,000 dollars, the second place gets 200,000 dollars, and the third place gets 150,000 dollars. The top three will get a total of 650,000 dollars.}
\end{quote}

This response showcases Breeze2 8B's ability to accurately understand the user's question in Traditional Chinese and extract the relevant information from the image. The model not only identifies the prize amounts for the top three positions but also performs the necessary calculation to provide the total sum. This example illustrates Breeze2 8B's capability to seamlessly integrate visual and textual information, demonstrating its potential for handling complex, multimodal tasks in Traditional Chinese.

\end{CJK*}

\section{Mobile APP}
\label{sec:app}

\textbf{Motivation and Features of the App.} To showcase the capabilities of \textit{Breeze2} 3B in a practical, user-friendly format and demonstrate its potential for real-world applications, we developed a mobile application tailored for Taiwan-localized knowledge and multi-modal interactions. The app, powered by the \textit{Breeze2} 3B model, excels in understanding Taiwan's cultural, historical, and geographical contexts in Traditional Chinese. With vision-aware capabilities, the app recognizes and interprets visual content such as landmarks, cultural artifacts, and signage, providing detailed explanations and insights. By supporting multi-modal inputs, users can combine text and images, enabling comprehensive descriptions, translations, and historical information, enhancing the experience for both local and international users.

\begin{figure}[!ht]
    \centering
    \includegraphics[width=0.9\textwidth]{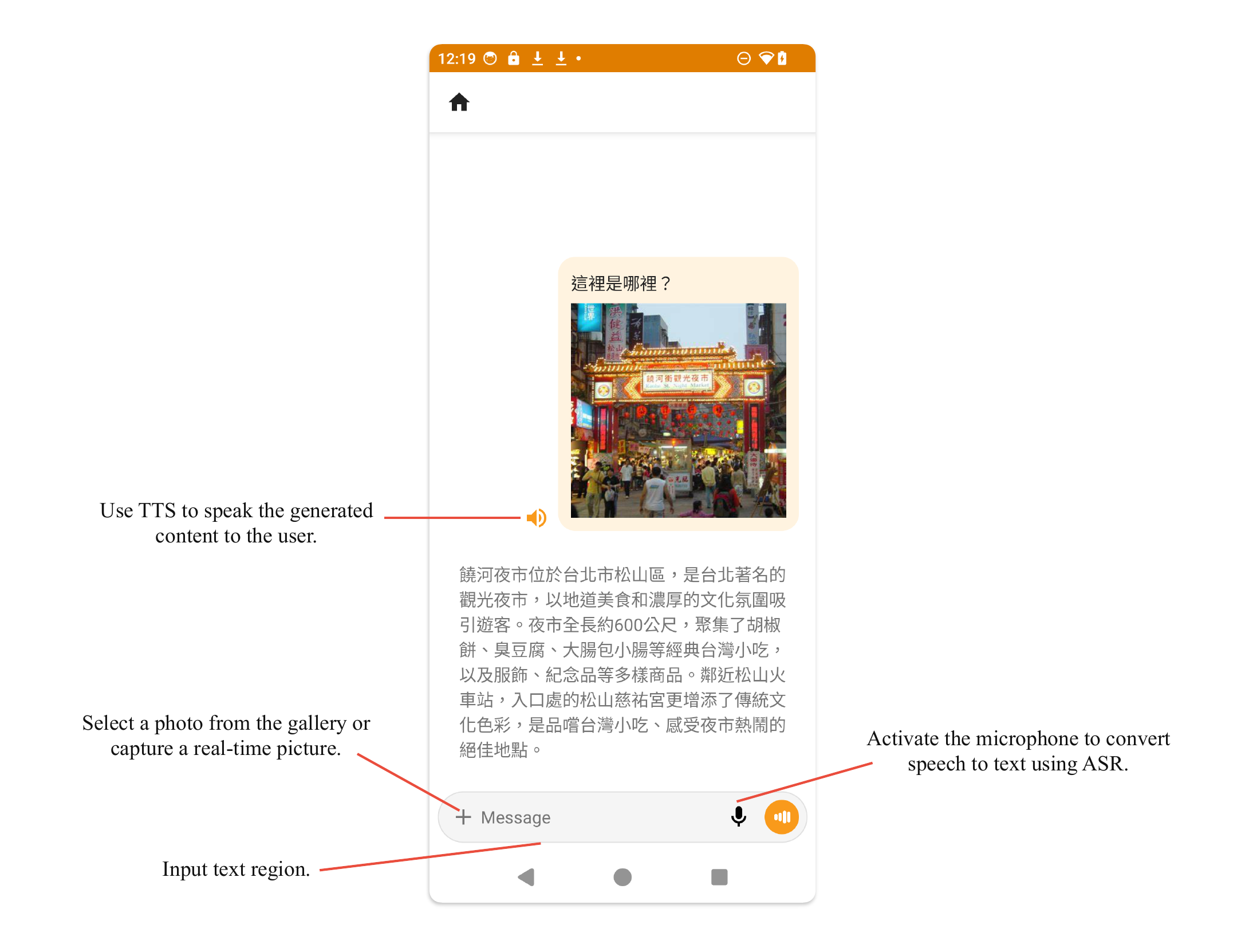} 
    \caption{Overview of the App Interface Design. This figure illustrates the main user interaction elements in the application. Users can select a photo from the gallery or capture a real-time picture to initiate visual processing. A dedicated input text region enables users to provide textual commands. Additionally, the application integrates a microphone for automatic speech recognition (ASR) to convert speech into text, allowing for hands-free interaction. The system also supports text-to-speech (TTS) functionality, which vocalizes generated content back to the user, ensuring accessibility and an enhanced user experience.}
    \label{fig:app_interface}
\end{figure}

\textbf{Implementation.} The \textit{Breeze2} 3B model, which integrates both language and vision capabilities, is integrated into the mobile application using ExecuTorch\footnote{\url{https://pytorch.org/executorch/stable/index.html}}, a tool designed for preparing and deploying large-scale models on Android devices. For the purposes of this implementation, we utilize two configurations of the \textit{Breeze2} 3B model: a language-only version (referred to as LLM in this context) based on the Llama 3.2 3B architecture, and the full vision-language integrated version (referred to as VLM). ExecuTorch converts these models into .pte files and tokenizers into .bin files, optimized for on-device inference. The ExecuTorch API, accessed via a JNI layer, supports tasks such as model initialization, parameter adjustments, and inferencing with delegates like XNNPACK (CPU-based) and MediaTek AI Accelerators.

Developed in Android Studio\footnote{\url{https://developer.android.com/studio?hl=zh-tw}}, the app handles both LLM and VLM functionalities, enabling users to load models, adjust parameters, and input system or user prompts. For vision-aware tasks, users can upload images or take live pictures, combining them with textual prompts for multi-modal interactions.

The app has been tested with selected configurations, focusing on the language-only and full vision-language integrated versions of the \textit{Breeze2} 3B model. Initial results demonstrate that ExecuTorch effectively supports multi-modal interactions, confirming its suitability for deploying advanced AI models on Android platforms.

\textbf{Performance.} 
The Android application was developed and tested on a smartphone equipped with the MediaTek Dimensity 9400 chipset. Although this platform is equipped with a high-performance AI accelerator (NPU), we opted to use the CPU-based XNNPACK delegate for inference. This choice ensures a fair comparison of performance metrics across devices, as XNNPACK is a widely supported and standardized delegate for on-device AI computations.

Key performance metrics, including model loading time, memory usage, prefill rate (tokens per second), and text decoder rate, are summarized in \autoref{tab:android_performance} for the language-only configuration of \textit{Breeze2} 3B.

\begin{table}[h!]
\centering
\caption{Performance Metrics on Android Device with XNNPACK delegate for \textit{Breeze2} 3B (language-only configuration)}
\label{tab:android_performance}
\begin{tabular}{l|r}
\hline
Performance Metric & Value \\
\hline
Model Loading Time (s)          & 2.48         \\
Memory Usage (GB)               & 6.87         \\
Prefill Rate (Tokens/s)   & 17.07        \\
Text Decoder Rate (Tokens/s)    & 3.83         \\ \hline
\end{tabular}
\end{table}

Note: The performance metrics for the full vision-language integrated version of \textit{Breeze2} 3B are currently being evaluated and will be reported in future updates.

\textbf{Limitations.} While the app demonstrates significant strengths, it does face some limitations. Despite the use of quantization, the 3B model remains resource-intensive, posing challenges for older devices with limited RAM and storage. Vision-aware functionalities, such as processing large image inputs or performing real-time visual analysis, can experience latency on mid-range devices, which may affect the responsiveness of the application. The app’s memory requirements are relatively high, which could impact multitasking capabilities, especially on devices with constrained resources.

\section{Conclusion and Future Work}

In this work, we introduced \textit{Breeze2}, a suite of multi-modal language models designed to address the underrepresentation of Traditional Chinese in large language models. By building upon the knowledge base of Llama 3 and incorporating extensive pre-training on Traditional Chinese corpora, we have developed models that excel in understanding and generating content relevant to Taiwan's linguistic and cultural context.

Our comprehensive multi-stage training approach, which includes extended text-to-text pre-training, vision-alignment pre-training, and post-training for instruction tuning and function calling, has resulted in models that demonstrate strong performance across a wide range of tasks. The evaluation results on various benchmarks highlight \textit{Breeze2}'s capabilities in general knowledge, instruction following, long context handling, function calling, and vision understanding, particularly in Traditional Chinese contexts.

Key achievements of \textit{Breeze2} include:

\begin{itemize}
    \item Improved performance on Taiwan-specific knowledge tasks, as evidenced by the results on TMMLU+ and MT-Bench-tw.
    \item Preservation of long-context capabilities inherited from Llama 3, with both 3B and 8B models showing high accuracy in passkey retrieval tasks across 128k token contexts.
    \item Competitive function calling abilities in both English and Traditional Chinese, comparable to the commercial model GPT-4o-mini.
    \item Promising results in vision-language tasks, our \textit{Breeze2} 8B outperforms the larger Llama-3.2 11B model across both English and Traditional Chinese benchmarks.
\end{itemize}

While \textit{Breeze2} represents an advancement in addressing the needs of Traditional Chinese language processing and Taiwan-specific applications, there are several avenues for future work:

\begin{enumerate}
    \item \textbf{Further enhancement of vision-language capabilities:} While \textit{Breeze2} shows promising results on vision-language tasks, there is room for improvement, especially when compared to larger models like GPT-4o-mini. Future work could focus on refining the vision-alignment training process and expanding the visual instruction tuning dataset.
    
    \item \textbf{Optimization for mobile deployment:} Given the successful implementation of \textit{Breeze2} 3B in a mobile application, future work could focus on further optimizing the model for mobile environments, potentially exploring techniques like quantization and model compression to improve efficiency without sacrificing performance.
    
    \item \textbf{Exploration of larger model sizes:} While the current 3B and 8B models show strong performance, investigating the potential benefits of larger model sizes for Traditional Chinese processing could yield interesting insights and potentially even better results.
\end{enumerate}

In conclusion, \textit{Breeze2} addresses the underrepresentation of Traditional Chinese in LLMs and incorporates vision-aware and function-calling capabilities. By focusing on Taiwan-specific applications, \textit{Breeze2} contributes to the development of more inclusive and culturally aware AI applications. As we continue to refine and expand upon this work, we anticipate \textit{Breeze2} will play a role in bridging the gap between advanced AI technologies and the unique linguistic and cultural needs of Traditional Chinese-speaking communities.

\section*{Author Contributions}
\begin{CJK*}{UTF8}{bsmi}
We list all contributors in alphabetical order of their first names, as follows: Chan-Jan Hsu (許湛然), Chia-Sheng Liu (劉佳昇), Meng-Hsi Chen (陳孟羲), Muxi Chen (陳沐希), Po-Chun Hsu (許博竣), Yi-Chang Chen (陳宜昌), and the supervisor Da-Shan Shiu (許大山).

Po-Chun Hsu and Yi-Chang Chen are the project leads of this work. Chan-Jan Hsu and Po-Chun Hsu contributed to the textual pre-training data. Po-Chun Hsu and Yi-Chang Chen contributed to the pre-training and post-training recipes. Yi-Chang Chen and Muxi Chen contributed to the experiments of function calling cabaility. Chia-Sheng Liu, Meng-Hsi Chen, and Yi-Chang Chen contributed to the experiments of vision-aware capability. Muxi Chen contributed to the mobile app.
\end{CJK*}

\newpage

\bibliographystyle{plainnat}
\bibliography{reference}

\end{document}